\documentclass[journal,twoside,web]{ieeecolor}
\usepackage[table]{xcolor} % Include xcolor with table option
\usepackage{tmi}
\usepackage{cite}
\usepackage{colortbl} % Include colortbl package
\usepackage{tabularx}
\usepackage{makecell} % for table
\usepackage{url}
\usepackage{float}
\usepackage{booktabs} 
\usepackage{amsmath,amssymb,amsfonts}
\usepackage{algorithmic}
\usepackage{graphicx}
\usepackage{textcomp}
\usepackage{balance}

\def\BibTeX{{\rm B\kern-.05em{\sc i\kern-.025em b}\kern-.08em
    T\kern-.1667em\lower.7ex\hbox{E}\kern-.125emX}}
{Alwazzan \MakeLowercase{\textit{et al.}}: Multimodal Outer Arithmetic Block Dual Fusion}

\usepackage{etoolbox}
\makeatletter
\@ifundefined{color@begingroup}%
  {\let\color@begingroup\relax
   \let\color@endgroup\relax}{}%
\def\fix@ieeecolor@hbox#1{%
  \hbox{\color@begingroup#1\color@endgroup}}
\patchcmd\@makecaption{\hbox}{\fix@ieeecolor@hbox}{}{\FAILED}
\patchcmd\@makecaption{\hbox}{\fix@ieeecolor@hbox}{}{\FAILED}

\begin{document}
\title{Multimodal Outer Arithmetic Block Dual Fusion of Whole Slide Images and Omics Data for Precision Oncology}
\author{Omnia Alwazzan, Amaya Gallagher-Syed, Thomas O. Millner, Sebastian Brandner, Ioannis Patras, \IEEEmembership{Senior Member, IEEE}, Silvia Marino, Gregory Slabaugh, \IEEEmembership{Senior Member, IEEE}
\thanks{The authors appreciate the support of the University of Jeddah and the Saudi Arabia Cultural Bureau. This paper utilised Queen Mary's Andrena HPC facility. This work also acknowledges the support of the National Institute for Health and Care Research Barts Biomedical Research Centre (NIHR203330), a delivery partnership of Barts Health NHS Trust, Queen Mary University of London, St George’s University Hospitals NHS Foundation Trust and St George’s University of London. }
\thanks{The authors are affiliated with Queen Mary University of London.  Corresponding author (e-mail:o.alwazzan@qmul.ac.uk). }
\thanks{Sebastian Brandner is with the Division of Neuropathology, The National Hospital for Neurology and Neurosurgery, University College London Hospitals NHS Foundation Trust, and Department of Neurodegenerative Disease, Queen Square, Institute of Neurology, University College London, London, United Kingdom (e-mail: s.brandner@ucl.ac.uk).}
% \thanks{T. C. Author is with the Electrical Engineering Department, University of Colorado, Boulder, CO 80309 USA, on leave from the National Research Institute for Metals, Tsukuba, Japan (e-mail: author@nrim.go.jp).}
}

\maketitle

\begin{abstract}
The integration of DNA methylation data with a Whole Slide Image (WSI) offers significant potential for enhancing the diagnostic precision of central nervous system (CNS) tumor classification in neuropathology. While existing approaches typically integrate encoded omic data with histology at either an early or late fusion stage, the potential of reintroducing omic data through dual fusion remains unexplored. In this paper, we propose the use of omic embeddings during early and late fusion to capture complementary information from local (patch-level) to global (slide-level) interactions, boosting performance through multimodal integration. In the early fusion stage, omic embeddings are projected onto WSI patches in latent-space, which generates embeddings that encapsulate per-patch molecular and morphological insights. This effectively incorporates omic information into the spatial representation of the WSI. These embeddings are then refined with a Multiple Instance Learning gated attention mechanism which attends to diagnostic patches. In the late fusion stage, we reintroduce the omic data by fusing it with slide-level omic-WSI embeddings using a Multimodal Outer Arithmetic Block (MOAB), which richly intermingles features from both modalities, capturing their correlations and complementarity. We demonstrate accurate CNS tumor subtyping across 20 fine-grained subtypes and validate our approach on benchmark datasets, achieving improved survival prediction on TCGA-BLCA and competitive performance on TCGA-BRCA compared to state-of-the-art methods. This dual fusion strategy enhances interpretability and classification performance, highlighting its potential for clinical diagnostics.

%Keep the abstract to 250 words or less.
\end{abstract}


\section{Introduction}
\label{sec:introduction}
%\IEEEPARstart{D}{NA}  methylation is an epigenetic process where methyl groups are added to specific sites on DNA molecules, typically at cytosine bases followed by guanine (CpG sites) \cite{lopomo2018epigenetic}. These modifications affect downstream gene expression levels without changing the DNA sequence itself, resulting in heritable changes in gene activity—an effect termed ``epigenetic''~\cite{tramacere2024neuro}. Such a process often leads to gene silencing or reduced gene expression by preventing transcription factors from binding to DNA, thereby inhibiting gene activation.
\IEEEPARstart{E}{pigenetics} describes a raft of molecular mechanisms which affect gene expression without changing the DNA sequence itself. DNA methylation is an epigenetic process where methyl groups are added to specific sites on DNA molecules, typically at cytosine bases followed by guanine (CpG sites) \cite{lopomo2018epigenetic}.  Such a process often leads to gene silencing or reduced gene expression by preventing transcription factors from binding to DNA, thereby inhibiting gene activation, but affects gene expression in many complex ways. The DNA methylation landscape of a cell is dependent on its developmental course and function, and this landscape can be grossly disrupted in the setting of cancer. In central nervous system (CNS) tumors, DNA methylation patterns offer valuable insights, enabling the differentiation of tumor subtypes, prediction of clinical outcomes, and guidance on treatment strategies \cite{liang2023integrative}. Diagnosis solely based on the Whole Slide Image (WSI) can be challenging due to the overlapping appearance of different tumor subtypes, resulting in high inter-observer variability \cite{drexler2024unclassifiable}. 

Given the aggressive nature of malignant CNS tumors and their associated poor survival rates, there is a critical need to improve diagnostic precision \cite{djirackor2021intraoperative}, expedite diagnostic timeframes and identify targets for future personalized treatments. Importantly, evaluation of digitized WSI is becoming increasingly utilized in these clinical diagnostic pathways. Incorporating DNA methylation profiling for diagnosis, one study in pediatric patients demonstrated altered subtype classifications in 35\% of cases, potentially impacting treatment decisions for 4\% of pediatric patients \cite{pickles2020dna},  while another study in an adult population showed diagnosis was changed in 25\%, refined in 4\% and confirmed in  25\% of cases \cite{jaunmuktane2019methylation}. This demand has driven the development of CNS tumor classifiers based on DNA methylation array data, providing significant advancements in neuropathology \cite{drexler2024unclassifiable}. The World Health Organization (WHO) has also responded by incorporating molecular profiling alongside traditional histology into the latest CNS tumor classification guidelines, which defines 40 tumor types and subtypes based on key molecular characteristics \cite{smith2022major}. Such findings have inspired further research into automated integrative molecular morphology classification systems using artificial intelligence (AI) algorithms, including machine learning (ML) and deep learning (DL) approaches, to improve tumor diagnosis and prognosis  \cite{chen2022pan,capper2018dna,chen2021multimodal,chen2024towards,drexler2024unclassifiable,hoang2024prediction,jaume2024modeling}.

\textbf{Unimodal Approaches.} In the single modality domain, Capper et al. \cite{capper2018dna} pioneered a methylation-based classification system on 2801 CNS tumor samples using a ML approach \cite{capper2018dna}. For each CNS tumor subtype included in the classifier, also known as the methylation class, the classifier generated a predicted probability (calibrated score) that summed to 1 \cite{capper2018practical}, with tumors with a score below 0.3 classified as ``no match''. Capper's method now serves as an essential aid in the routine diagnostic workup of CNS tumors \cite{drexler2024unclassifiable}. Hwang et al. \cite{hwang2024image} have developed an image-based DL model using DNA methylation data to predict the origin of cancers of unknown primary (CUPs). By employing a vision transformer to organ-specific DNA methylation images, their approach shows significant potential for enhancing CUP diagnosis and informing treatment strategies. DNA methylation profiling not only aids in precise classification but also supports surgical strategies specific to CNS tumors, improving surgical outcomes and overall patient care \cite{djirackor2021intraoperative}. Djirackor et al. \cite{djirackor2021intraoperative} utilize ML algorithms to classify brain tumors in real-time by taking the DNA methylation data of a new tumor sample and comparing it to a database of known methylation signatures, assigning a classification based on the closest match. This allows for fast intraoperative decision-making by providing molecular insights during surgery.

\textbf{Multimodal Approaches.} In the multimodal domain, Hoang et al. \cite{hoang2024prediction} developed ``Deploy'', a DL model designed to predict DNA methylation beta values from WSIs. Additionally, Zheng et al. \cite{zheng2020whole} demonstrated that classical ML algorithms can link DNA methylation profiles of cancer samples with morphometric features from WSIs, showing improved model performance when genes are grouped into methylation clusters. Sturm et al. \cite{sturm2023multiomic} explore the use of a multiomic approach — integrating genomics, transcriptomics, and epigenomics data — to improve the diagnostic accuracy of pediatric brain tumors, which are often challenging to classify due to overlapping histological features. However, few studies \cite{hoang2024prediction} have combined epigenetic data with WSIs, primarily due to the significant data size and complexity of both modalities, which require extensive preprocessing and developing advanced fusion techniques to address their heterogeneity. 

Multimodal DL approaches combining histology and omic data for improved survival prediction have gained considerable attention in recent years \cite{chen2022pan, chen2021multimodal, chen2024towards, jaume2024modeling, mobadersany2018predicting, zhang2024mbfusion, zhang2024prototypical, ogundipe2024deep, song2024multimodal, zhang2023attention, ramanathan2024ensemble, liu2024agnostic}. Several studies \cite{song2024multimodal, jaume2024modeling, xu2023multimodal, chen2021multimodal} have highlighted the value of different fusion stages (early or late), particularly emphasizing early fusion for its ability to create an explainable framework from heterogeneous data. However, Zhang et al. \cite{zhang2024prototypical} argue that early or late fusion methods can partially overlook modality-specific information, potentially leading to a decline in quantitative or qualitative performance. Accordingly, they propose a prototypical information bottleneck framework to maintain modality-specific information while simultaneously reducing redundancy. Furthermore, a joint distribution of feature embeddings is used to calculate the mutual information between omic and WSI modalities. 

Despite the promising performance of multimodal approaches in medical diagnostics, significant challenges remain in effectively integrating and analyzing diverse data types, particularly in the context of CNS tumor subtyping. Our assessment of existing methods reveals three key gaps: \\ \\

1) \textbf{Limitations of WSI-only diagnosis}: While WSI serves as a primary diagnostic tool for pathologists, accurately identifying fine-grained tumor subtypes based on morphology alone is challenging due to visual feature overlap among CNS subtypes. This similarity increases the risk of including irrelevant tumor regions and highlights the need for complementary data sources to achieve more precise subtyping \cite{drexler2024unclassifiable}.

2) \textbf{Shortcomings of DNA methylation classification}: DNA methylation classifiers have demonstrated high accuracy in tumor subtyping, but they lack the ability to connect this accuracy to specific regions within WSIs. The main limitation is that these classifiers focus solely on DNA methylation profiles without considering the spatial context provided by WSIs. This limits their capacity to capture how epigenetic patterns contribute to the morphological characteristics observed in specific regions, ultimately reducing the interpretability and comprehensive understanding that could be achieved through integration with WSI data.

3) \textbf{Scarcity of integrative models for CNS tumor subtyping}: This challenge is particularly acute in the context of central nervous system tumors, where the inherent heterogeneity and large size of both DNA methylation arrays (typically 850k one-dimensional vectors) and WSIs (represented with multi-dimensional matrices up to 150k x 150k pixels) have received limited attention. The lack of effective multimodal fusion methods in this domain presents an opportunity to leverage the powerful discriminative capacity of DNA methylation data to improve subtyping and enhance the clinical translation of these advanced imaging and molecular techniques. 

% Addressing these gaps requires innovative approaches to data fusion, potentially exploring both early and late fusion strategies, to fully leverage the complementary strengths of WSI and DNA methylation data in CNS tumor subtyping.

%In this study, we present a novel approach to enhance the accuracy of CNS tumor subtyping by integrating DNA methylation profiling with histological features. We employed a dual fusion strategy, introducing DNA methylation data at two stages: early fusion to improve interpretability and late fusion to enhance classification accuracy, thereby maximizing the diagnostic potential of DNA profiles. This method ultimately provides a more precise and comprehensive classification of CNS tumors.
\subsection{Contributions}
Addressing these challenges, we propose a novel dual fusion approach to improve CNS tumor subtyping by integrating DNA methylation data with WSIs. We design MOAD-FNet, a \textbf{M}ultimodal \textbf{O}uter \textbf{A}rithmetic \textbf{D}ual \textbf{F}usion \textbf{Net}work that combines two fusion variants: early fusion focused on capturing essential local interactions and late fusion for broader, richer cross-modal global context, ultimately providing complementary insights and improving the model's decision-making process. This dual fusion strategy provides a comprehensive, holistic integration of cross-modal data, maximizing the strengths of each fusion type. Our main contributions are as follows:
\begin{itemize}
    \item We introduce a novel dual fusion network that seamlessly incorporates both early and late fusion approaches, enabling detailed integration of molecular and imaging data at patch and slide levels.
    \item Our approach develops a unique Multimodal Outer Arithmetic Block (MOAB) fusion strategy that enhances cross-modal feature interaction and improves the model's ability to capture complex tumor subtype features.
    \item Our method, MOAD-FNet, is the first imaging-omics pipeline to leverage the NHNN BRAIN UK dataset, demonstrating exceptional performance in brain tumor subtyping. Extensive ablation studies on TCGA datasets validate its robustness, achieving state-of-the-art survival prediction on the BLCA dataset and on par results on the BRCA dataset, showcasing its versatility across multimodal oncology tasks.
\end{itemize}
We argue that using a late fusion variant alone would capture only global interactions between the WSI level label and epigenetic data, lacking the interpretability needed for accurate subtyping \cite{jaume2024modeling, song2024morphological}. On the other hand, an early fusion approach alone would focus exclusively on local interactions, neglecting the broader relationships between modalities \cite{song2024morphological}, which we confirm with extensive experimentation. Our proposed dual fusion framework, MOAD-FNet, addresses these limitations by combining early and late fusion strategies, creating a new approach that uniquely captures both local and global interactions across modalities and enhances interpretability and overall classification accuracy.

\begin{figure*}[ht]
\centering
\includegraphics[width=\textwidth,height=0.5\textheight,keepaspectratio]{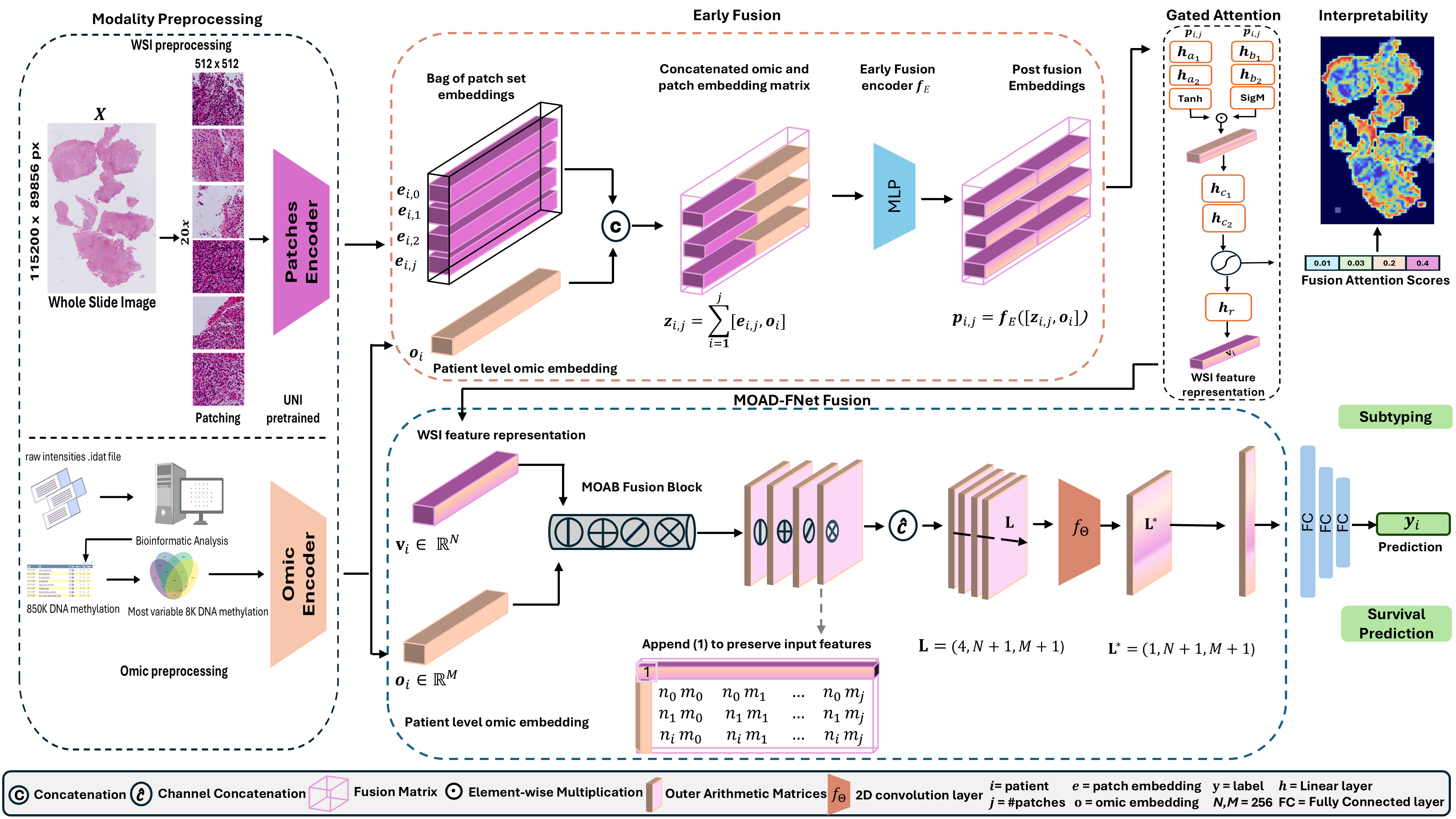}

\caption{Overview of the proposed MOAD-FNet framework. Data engineering and encoding for each modality are performed in the preprocessing block. The early fusion block (top) receives encoded inputs from both modalities, where omic data is concatenated to form a matrix $\mathbf{z}_{i,j}$ which is processed by an MLP encoder that learns a joint mapping, resulting in output $\mathbf{p}_{i,j}$. A gated attention via multiple instance learning (MIL) scores patch importance providing heatmap interpretability and producing a WSI feature $\mathbf{v}_i$. Next, the MOAD-FNet fusion block (bottom) reintroduces omic features $\mathbf{o}_i$ alongside the $\mathbf{v}_i$ feature representation from the early fusion block as input to the MOAB fusion block. This block performs four outer arithmetic operations to create fusion representations, which are further reduced with $f_\theta$ before being sent to the final subtyping classifier.}
\label{fig1}
\end{figure*}

Note that, this paper extends our previous conference paper~\cite{alwazzan2023moab} by implementing a dual fusion architecture,  working with three additional imaging-omics datasets, recently proposed state-of-the-art (SOTA) backbones, and shows how heatmaps can be generated, providing interpretability.
\section{Study Design}

\subsection{Datasets}

The following sections briefly provide an overview of the datasets used to evaluate MOAD-FNet.

\begin{figure}[H]
\centering
\includegraphics[width=0.9\columnwidth,height=0.18\textheight]{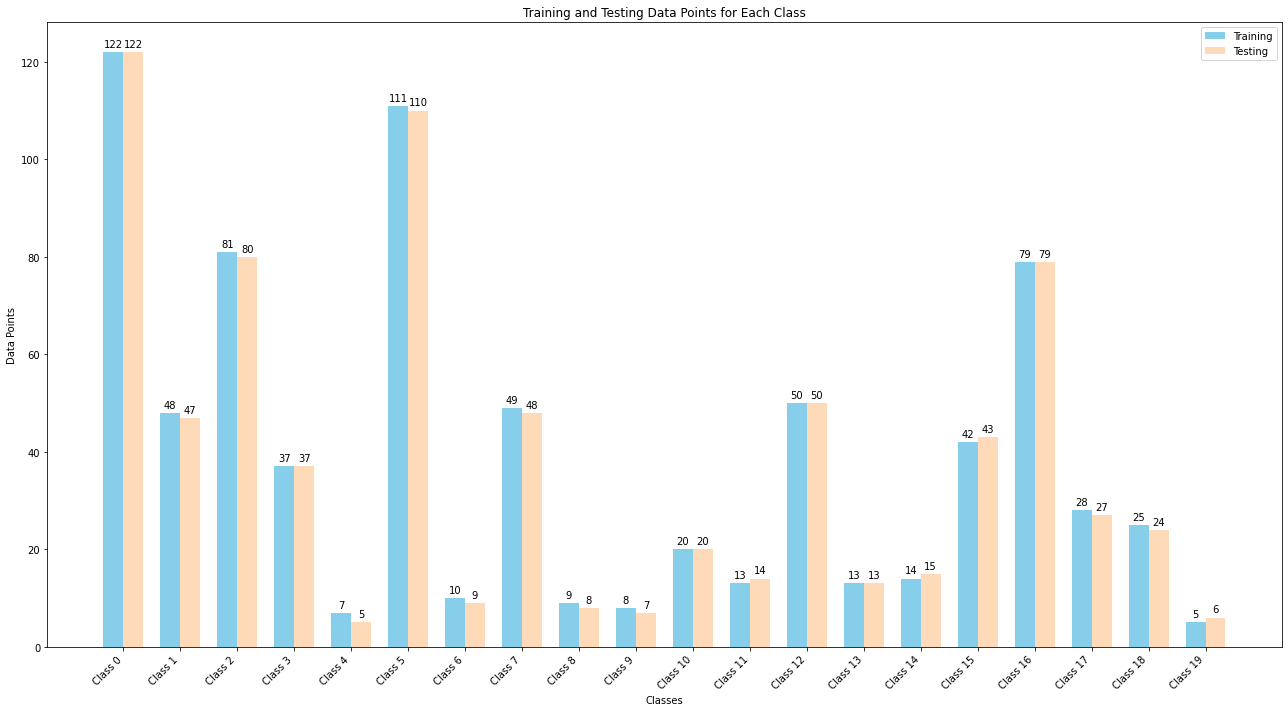} 
\caption{Distribution of training and testing data points across 20 classes/subtypes. The bar chart illustrates the number of patients allocated to training and testing sets for each class, highlighting the balance of data used for model development and evaluation.}
\label{fig2}
\end{figure}

\subsubsection{Slide Subtyping - NHNN BRAIN UK Brain Tumor Dataset}
We obtained WSIs and matched DNA methylation data from the NHNN, University College London Hospital (UCLH) through the UK Brain Archive Information Network \cite{nicoll2022brain}\footnote{https://www.southampton.ac.uk/brainuk}.
%We obtained WSIs and matched DNA methylation data from the UK Brain Archive Information Network (BRAIN UK)\cite{nicoll2022brain}\footnote{https://www.southampton.ac.uk/brainuk/}. 
The dataset includes (H\&E)-stained WSIs from 1,504 patients, covering 20 DNA-methylation-based subtypes of high- and low-grade glial tumors. The data exhibits significant class imbalance across its 20 subtypes, as shown in Fig. \ref{fig2}. To ensure a balanced evaluation, we perform 2-fold cross-validation by splitting the dataset into two equal parts (50\%-50\%). In each fold, one part is used for training and the other for testing, ensuring all slides are evaluated while maintaining a strict separation between the two sets. We report the average performance across both folds for all metrics. For WSIs, we used QuPath \cite{qupath} for tissue segmentation and tiling, generating 1M patches (4K–10K per slide).  For the tissue methylation profile, we processed raw idat files to extract DNA methylation CpG sites using the hg38 Illumina array with EPICv1 annotation, following the workflow in \cite{orozco2018epigenetic}. We calculated m-Values to quantify methylation levels, producing a matrix of 850k CpG sites for 1,504 patients. To reduce features, we applied variance, coefficient of variation (CV), median absolute deviation, and inter-quartile range, selecting the 8K most variable CpG sites by intersecting these methods. We experimented with clustering 4K CpG sites and Random Forest selection of 10K sites, as suggested in the literature\cite{capper2018dna,orozco2018epigenetic,gomes2022application}, but 4K reduced performance and 10K showed no improvement, leading us to select 8K CpG sites.%Unlike standard methods \cite{lu2021data} that can produce up to 300K patches per slide—posing significant computational challenges—we opted not to use these approaches. The large size of the UK brain images contributed to these issues, and QuPath effectively addressed them by providing reasonable resolution patches with good coverage, yielding a manageable number of patches per slide (ranging from 4K to 10K). 

\subsubsection{Survival prediction - TCGA Datasets}
To showcase MOAD-FNet's versatility and compare it with other omic-WSI fusion methods \cite{chen2021multimodal, chen2022pan, jaume2024modeling, shao2021transmil, song2024multimodal}, we evaluate its performance on The Cancer Genome Atlas (TCGA) datasets for Bladder Urothelial Carcinoma (BLCA) $(n=359)$ and Breast Invasive Carcinoma (BRCA) $(n=869)$, using the survival prediction task outlined by Jaume et al. \cite{jaume2024modeling}. We utilized a coupled set of 331 biological pathways derived from 4,999 distinct genes. These genes were grouped into pathways based on their functional interactions and roles in biological processes relevant to the BLCA and BRCA datasets.

\subsection{Multimodal Outer Arithmetic Dual Fusion Network}
We designed a multimodal fusion framework, shown in Fig.~\ref{fig1}, that integrates omic data and a WSI through combined early and late fusion stages. Our proposed method, MOAD-FNet, is aimed at multimodal brain tumor subtyping, and survival prediction in both lung and breast cancer. In the following subsections, we highlight key components of the MOAD-FNet framework. 

\subsubsection{Omic Encoder}\label{omic-encoder}
To construct the omic encoder, we followed the established practice of using a Self-Normalizing Neural (SNN) \cite{klambauer2017self} consisting of two fully connected layers, where each layer applies an Exponential Linear Unit (ELU) activation function followed by Alpha Dropout (0.25). The SNN compresses the 8K CpG sites into an encoded omic feature ${\mathbf{o}_i}\in \mathbb{R}^{d^o}$ ${d^o}=256$. Note that for fair comparison with other methods \cite{song2024multimodal, jaume2024modeling, zhang2024prototypical, mobadersany2018predicting}, we used the same encoder tokenizing the genes into a group of pathways.
\subsubsection{Whole Slide Image Encoder}
For a WSI $\mathbf{X}$, we extract a collection of patches, represented as ${\mathbf{x}_{i0}, \mathbf{x}_{i1},\ldots, \mathbf{x}_{ij}}$, where $i$ represents the patient/slide index and $j$ indicates the patch index that varies across slides. We extract non-overlapping patches from tissue areas at a 20× magnification (about 0.5 µm/pixel resolution). Subsequently, we utilized a SOTA image-only encoder \cite{chen2024towards} UNI to obtain the patch embeddings ${\mathbf{e}_{ij}}$. Using UNI ${f_{enc}(.)}$, we derive a set of low-dimensional patch embeddings for each patient, where ${\mathbf{e}_{ij}} = {f_{enc}}(\mathbf{x}_{ij}) \in \mathbb{R}^{d^e}$, ${d^e}=1024$, serving as input to our pipeline.

\subsubsection{Fusion Stages}
The fusion scheme is the key contribution of this work. Our motivation sparks from the enhanced features obtained from the dense modeling conducted by multimodal early fusion methods in \cite{jaume2024modeling, song2024multimodal, chen2021multimodal, xu2023multimodal}. Hence, we divide this subsection into two: early fusion and late fusion.

\textbf{Early Fusion.} Given a matrix of patch embeddings ${\mathbf{e}_{ij}}$ and the omic feature vector ${\mathbf{o}_i}$ for patient $i$, the encoded omic feature vector is cloned to match the number of patches in the WSI. This results in a tensor of shape $(N_i, d_o)$, where $N_i$ is the number of patches in WSI $i$, and $d_o$ is the dimension of the omic feature. The combined feature set for WSI 
$i$ is represented as:
     \begin{equation}
     {\mathbf{z}_{ij}} = [{\mathbf{e}_{ij}}, {\mathbf{o}_i}]
\end{equation}
Here, ${\mathbf{z}_{ij}} \in \mathbb{R}^{d_e + d_o}$ represents the concatenated $[.]$ feature vector for patch $j$ in WSI $i$, and the resulting shape becomes $(N_i, d_e + d_o)$. Next, a Multilayer Perceptron (MLP) encoder, denoted as \( f_{E} \), is  applied to each concatenated patch embedding and omic feature pair to learn their joint representation:

\begin{equation}
\mathbf{p}_{ij} = f_{E}([\mathbf{z}_{ij}, \mathbf{o}_i])
\end{equation}

This operation is performed for all patches $j$ of the $i$th WSI. By incorporating omic features $\mathbf{o}_i$ into patch embeddings via early fusion, we enrich the representation with complementary molecular information. Leveraging an Attention-based Deep Multiple Instance Learning (ABMIL) approach\cite{ilse2018attention}, we capture patch-level discriminative features that synergistically combine molecular and morphological insights, enabling more precise identification of critical regions. The resulting embedding is projected to a slide-level representation $\mathbf{v}_i \in \mathbb{R}^{256}$, dimensionally aligned with the original omic feature $\mathbf{o}_i$ to facilitate subsequent late fusion. 

\textbf{Late Fusion}, motivated by \cite{jaume2024modeling} in modeling the interaction between omic to histology, histology to omic, and omic to omic, we mimic a similar behavior by employing our novel multimodal outer arithmetic fusion block (MOAB) \cite{alwazzan2023moab} within a dual fusion approach, marking a new direction in combining MOAB with dual fusion to enhance modality integration. MOAB inputs ($\mathbf{v}_i$,$\mathbf{o}_i$) will be fused through four operations: outer product, outer division, outer subtraction, and outer addition. MOAB extracts various interactions while simultaneously preserving the $\mathbf{v}_i$ input feature by appending one to each input embedding when performing the outer product and division fusion and zero in the case of the outer subtraction and addition fusion. To simplify this, given the two embeddings $\mathbf{v}_i$, $\mathbf{o}_i$,   $\mathbf{\mathbf{v}_i} \in \mathbb{R}^{N\times1}$ and $\mathbf{\mathbf{o}_i} \in \mathbb{R}^{M\times1}$. Initially, we append a 1 to each embedding, i.e., $\mathbf{v}_{i1} = [1; \mathbf{v}_i]$ and $\mathbf{o}_{j1} = [1; \mathbf{o}_j]$. Their outer product is defined as
\begin{equation}
    (\mathbf{\mathbf{v}}_{i1} \otimes \mathbf{o}_{j1})_{ij} = \mathbf{v}_{1i}*\mathbf{o}_{1j}, 
\end{equation}
for $i \in [1 ... N+1]$ and $j \in [1 ... M+1]$ yielding a $(N+1) \times (M+1)$ matrix that intermingles every element of $\mathbf{v}_{i1}$ with every element of $\mathbf{o}_{j1}$.  The appended 1 in both $\mathbf{v}_{i1}$ and $\mathbf{o}_{j1}$ ensures the original unimodal features $\mathbf{v}_i$ and $\mathbf{o}_j$  appear in the outer product matrix. Similarly, we define outer division, addition, and subtraction as
\begin{eqnarray}
    (\mathbf{\mathbf{v}}_{i0} \oplus \mathbf{o}_{j0})_{ij} & = & \mathbf{v}_{0i}+\mathbf{o}_{0j} \\
    (\mathbf{\mathbf{v}}_{i0} \ominus \mathbf{o}_{j0})_{ij} & = & \mathbf{v}_{0i}-\mathbf{o}_{0j} \\    
    (\mathbf{\mathbf{v}}_{i1} \oslash \mathbf{o}_{j1})_{ij} & = & \mathbf{v}_{1i}  \div ( \mathbf{o}_{1j} + \epsilon) 
\end{eqnarray}
where $\epsilon$ is a small number (set to $1\text{e}{-10}$), and $\mathbf{v}_{i0} = [0; \mathbf{v}_i]$ and $\mathbf{o}_{j0} = [0; \mathbf{o}_j]$.

The four matrices produced by MOAB are concatenated along the channel dimension to form a multimodal tensor $\mathbf{L}\in \mathbb{R}^{4\times257\times257}$. We hypothesize that channel fusion will maintain the proximity of closer points and will use fewer parameters compared to a typical concatenation. By combining features across the channel dimension, we greatly decrease dimension by compressing the feature representation from $(257\times257)^4$ to $(257)^2$. Following the same parameters in \cite{alwazzan2023moab}, a 2D convolution layer is subsequently performed to leverage associated interactions, resulting in a singular condensed multimodal feature tensor $\mathbf{L}^* \in \mathbb{R}^{1\times257\times257}$. Last, we flatten $\mathbf{L}^*$ and apply leaky ReLU followed by a linear predictor for brain tumor subtyping prediction. MOAB provides simple, yet effective operations to fuse multimodal data, similar to \cite{jaume2024modeling} but without relying on approximations.

\subsubsection{Interpretability}
Employing ABMIL \cite{mobadersany2018predicting} on the post-fusion embedding $\mathbf{p}_{ij}$ enables the visualization of attention scores, which provide an enriched perspective on the tumor's visual and omics characterization. This, in turn, enhances ABMIL’s capacity to prioritize patches that are biologically relevant. We define our gated attention computation as follows:
\begin{equation}
\begin{aligned}
    \mathbf{h}_{ij} &= \tanh(\mathbf{W}_p \mathbf{p}_{i,j} + \mathbf{b}_p) \odot \sigma(\mathbf{W}_g \mathbf{p}_{i,j} + \mathbf{b}_g), \\
    a_{ij} &= \frac{\exp(\mathbf{w}^T \mathbf{h}_{ij})}{\sum_{1}^j \exp(\mathbf{w}^T \mathbf{h}_{ij})}, \\
    \mathbf{v}_i &= f_\rho\left(\sum_{j=1}^j \mathbf{a}_{ij} \mathbf{h}_{ij}\right),
\end{aligned}
\end{equation}

\begin{figure*}[ht]
\centering
\includegraphics[height=0.2\textheight]{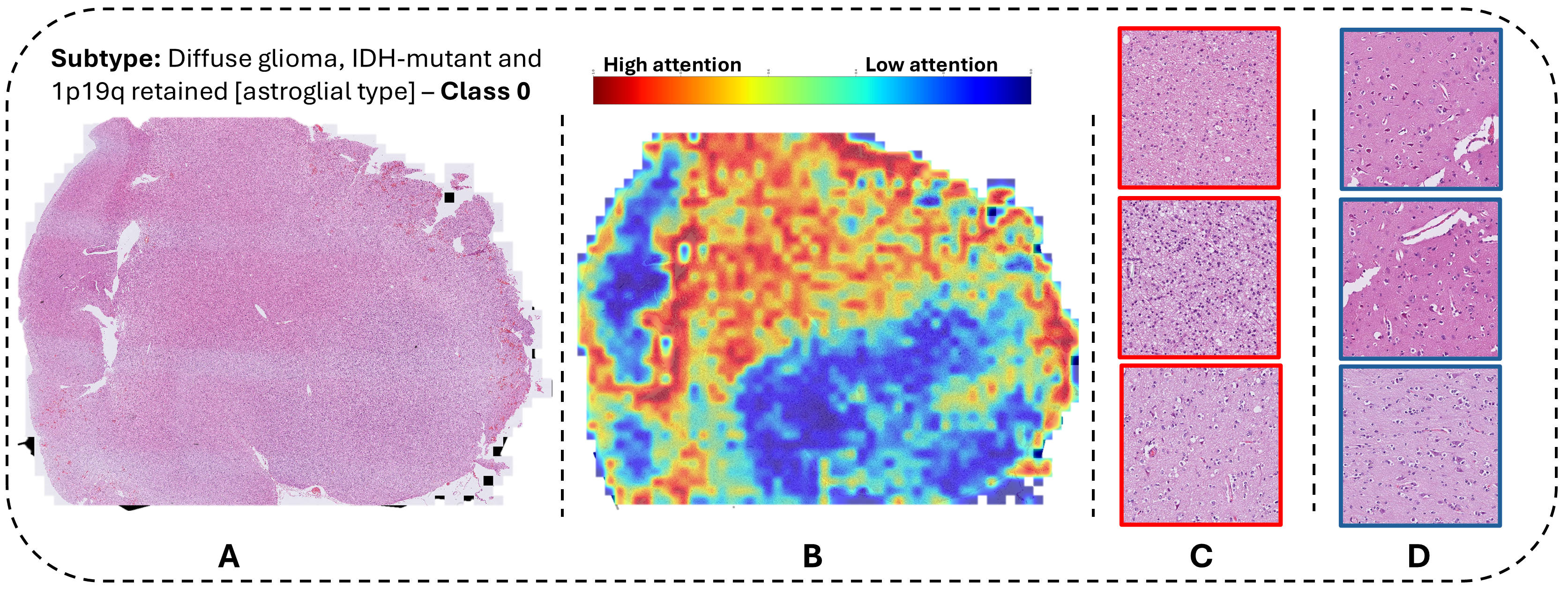} 
\caption{Visual representation of attention heatmap generated by MOAD-FNet for a diffuse glioma, IDH-mutant and 1p19q-retained (astroglial type) tumor (Class 0). (A) The original histology slide is displayed. (B) The heatmap shows areas of high attention (red) and low attention (blue), with regions of diagnostic relevance highlighted. (C) Representative patches with high attention are bordered in red, potentially indicating hallmark features of astroglial differentiation and cellular atypia crucial for diagnosis. (D) Representative patches with low attention are bordered in blue, reflecting regions of low tumor infiltration. The color bar illustrates the attention scale from high (red) to low (blue).}
\label{fig5}
\end{figure*}

Here, $\mathbf{W}_p$ and $\mathbf{W}_g$ are learnable parameters for the input and gating functions, respectively, with $\mathbf{b_p}$ and $\mathbf{b}_g$ as their corresponding biases.  The symbol $\odot$ denotes element-wise multiplication, and $\sigma$ represents the sigmoid activation function. The gated embedding for patch $j$ in slide $i$ is given by $\mathbf{h}_{ij}$, while $\mathbf{a}_{ij}$ represents the attention weight for patch $j$, normalized across all patches. We leverage $\mathbf{a}_{ij}$ at inference to generate the heatmap shown in Fig. \ref{fig5}. To further enhance our late fusion stage, we weight $\mathbf{v}_i$ with $\mathbf{a}_{ij}$ using an MLP $f_\rho$ which consists of a linear transformation, an activation function (ReLU), and dropout. $f_\rho$ is applied to the pooled embedding $\mathbf{v}_i$ increases its representational power, making it a more refined input for the late fusion stage. This enriched embedding encapsulates a high-level representation of the WSI, effectively integrating morphological and molecular insights from the attention mechanism.

%The corresponding unnormalized WSI-level score, \( s_{i,y} \), is given through the classifier layer \( \mathbf{V}_{y} \in \mathbb{R}^{1 \times 256} \) as:

%\begin{equation}
%s_{i,y} = \mathbf{V}_{y} W_{i,y}^\top
%\end{equation}

%For inference, the predicted probability distribution over each class is computed by applying a softmax function to the WSI-level prediction scores, \( s_i \). Similar to \cite{jaume2024modeling} \cite{song2024multimodal}, our early fusion approach facilitates the interpretation of the molecular interactions between complex DNA methylation data and the spatial regions of the whole slide images (WSI).% Furthermore, employing (MOAB) in the late fusion stage effectively differentiates rare classes from dominant ones, even when rare classes are underrepresented. This highlights its capability to capture complementary signals from both modalities, thereby enhancing model performance in a challenging subtyping scenario.
\section{Experiments}
%\subsection{Expert Annotations}
\subsection{Performance Metrics}

For the subtyping task, we assessed performance utilizing various metrics: F1-Macro, F1-Micro, Precision, and Recall/Sensitivity. F1-Macro is a significant metric for our quantitative analysis as it independently computes the F1 score for each class and subsequently averages them, assigning equal weight to each class irrespective of its size, thereby ensuring that the performance of minority classes is not overwhelmed by that of majority classes. For the survival prediction task, we follow the implementations of \cite{jaume2024modeling, zhang2024prototypical, song2024multimodal} where survival analysis is defined as an estimation of the probability of an event occurring within a given survival time, while accounting for right-censored data. Censorship status is represented as \( c \in \{0, 1\} \), where \( c = 0 \) denotes an observed event (e.g., death) and \( c = 1 \) indicates the patient's last known follow-up. In line with previous work we discretize the time-to-event into non-overlapping time intervals \( (t_{i-1}, t_i] \), based on the quartiles of survival times denoted as \( y_i \). This formulation transforms the problem into a classification task with censorship information, where each patient is represented by \( (\mathbf{L}_\text{logits}, y_i, c) \).

Next, we use the dual fusion embedding generated by MOAB-FNet, \( \mathbf{L}_\text{logits} \) to predict the discretized bin corresponding to a time interval \( t_i \). We define the discrete hazard function as:
\[
f_\text{hazard}(y_i \mid \mathbf{L}_\text{logits}) = \sigma(y_i),
\]
where \( \sigma \) is the sigmoid activation function. Intuitively, \( f_\text{hazard}(y_i \mid \mathbf{L}_\text{logits}) \) represents the probability that the patient experiences the event (e.g., death) during the interval \( (t_{i-1}, t_i] \). The discrete survival function is then defined as:
\[
f_\text{surv}(y_i \mid \mathbf{L}_\text{logits}) = \prod_{k=1}^{i-1} \left(1 - f_\text{hazard}(y_k \mid \mathbf{L}_\text{logits})\right),
\]
which represents the probability that the patient survives up to the interval \( (t_{i-1}, t_i] \). The Negative Log-Likelihood (NLL) survival loss is formally defined as:
\begin{multline}
   \mathcal{L}\left(\{\mathbf{L}_\text{logits}, y, c\}_{i=1}^{N_\text{total}}\right) = 
   - \sum_{i=1}^{N_\text{total}} \Big[ c_i \log\big(f_\text{surv}(y_i \mid \mathbf{L}_\text{logits})\big) \\
   + (1 - c_i) \log\big(f_\text{surv}(y_i \mid \mathbf{L}_\text{logits}) - 1[y_i]\big) \\
   + (1 - c_i) \log\big(f_\text{hazard}(y_i \mid \mathbf{L}_\text{logits})\big) \Big],
\end{multline}
where $( N_\text{total} )$ is the total number of samples in the dataset, and $( k )$ corresponds to the total number of discretized labels.
% As noted in\cite{jaume2024modeling}, this approach enhances survival analysis by accurately estimating survival probabilities, predicting event timestamps, and rigorously accounting for censorship status.
 
\subsection{Domain-Specific Prior Work and Ablation Studies}
To evaluate our method on subtyping and survival prediction tasks, we replicate and adapt recent SOTA methods by incorporating MOAB as a replacement for the fusion technique initially used in these methods. For the survival prediction task, we conducted a comparative analysis using a consistent feature extractor across all modalities, including WSI and omic data, utilising the recent TCGA ID samples from \cite{jaume2024modeling}. We uniformly implemented training hyperparameters and loss functions across all models displayed in Tables \ref{tab1} and \ref{tab3}.  To this end, this section is divided into two parts. SOTA baseline models detailing unimodal and fusion models employed for each task, together with a description of the ablation studies conducted.

\subsubsection{State-of-the-Art Baseline Models}
We evaluate our approach against SOTA techniques in the unimodal and multimodal setting for both subtyping and survival prediction tasks. \\

\textbf{Unimodal Subtyping baselines.}
We utilize MLP and SNN \cite{klambauer2017self} as baseline models for the DNA methylation data. For WSIs, we utilize Attention-based Multiple Instance Learning (ABMIL) \cite{mobadersany2018predicting}, which implements gated weighted attention pooling to determine the importance of patches, as well as Transformer-based Multiple Instance Learning (TransMIL) \cite{shao2021transmil} which employs the Nyström attention mechanism to evaluate correlations among WSI patches. Through rigorous experimentation, we found that ABMIL outperforms the TransMIL baseline in both performance and computational efficiency, also noted in \cite{jaume2024transcriptomics}, making it the preferred baseline for MOAD-FNet.

\textbf{Multimodal Subtyping baselines}.  
We adapt Attention Challenging Multiple Instance Learning (ACMIL) \cite{zhang2023attention} to work in a multimodal setting. ACMIL  is an approach designed to address overfitting in single-modality WSI classification, by using multiple attention branches and a composite loss (cross-entropy + diversity loss) to distribute attention across the WSI. We also used TransMIL with two late fusion variants, concatenation (Cat) and Kronecker product (KP) \cite{shao2021transmil}. Furthermore, we compare against MCAT \cite{chen2021multimodal} and SurvPath \cite{jaume2024modeling}, both of which perform multimodal tokenization to extract histology and biological pathway tokens. MCAT and SurvPath employ a transformer-based early fusion approach, the concatenate the resulting vectors. To maintain their architectural consistency, we opted not to replace concatenation with the MOAB late fusion block, as their complex architectures could be destabilized by it. However, to test our dual fusion hypothesis we input our early fusion input representation $\mathbf{p}_{ij}$ into MCAT and SurvPath, thus enabling these models to leverage omic-WSIs embeddings at both the early and late fusion stages. By comparing the performance of SurvPath and MCAT with $\mathbf{p}_{ij}$ to the original implementation (SurvPath* and MCAT*), we provide evidence that dual fusion outperforms single-stage fusion methods.

\textbf{Unimodal Survival baselines.}  
In addition to the unimodal subtyping baselines, we employ Sparse-MLP \cite{jaume2024modeling}, which tokenizes transcriptomics into biological pathway tokens encoding specific cellular functions for the downstream analysis.

\textbf{Multimodal Survival baselines.}  
We integrated MOAD-FNet across the same models used for subtyping tasks, adding the prototypical information bottlenecking and disentangling (PIBD) method \cite{zhang2024prototypical}. It introduces a disentanglement mechanism to separate modality-specific versus shared information. For PIBD, we restricted MOAB to a late fusion setting, respecting PIBD's initial modality-specific separation. \\

% Introducing MOAD-FNet in an early fusion setting could potentially conflict with this initial modality independence, underscoring our strategic late fusion choice to maintain alignment with PIBD’s architectural principles. \\

% To modify ACMIL for survival analysis, we integrated the NLL loss with the diversity loss to emulate the learning methodology employed in the original study \cite{zhang2023attention}.

\subsubsection{Ablation studies}
we conducted three ablation experiments to comprehensively evaluate MOAD-FNet. We first removed the MOAB fusion block, using only ABMIL. Here, $\mathbf{p}_{ij}$ served as the input to ABMIL, and the resulting feature embedding $\mathbf{v}_i$ was directly fed to the classifier layer to assess the distinct impact of both ABMIL and the DNA modality. In the second experiment, we removed the early fusion block (illustrated in Fig. \ref{fig1}), making $\mathbf{e}_{ij}$ the input to the gated attention block, which is then followed by MOAD-FNet. Last, we employed a task-agnostic encoder ConvNext.v2 \cite{liu2022convnet}, pre-trained on ImageNet, to extract features from the WSI and tested MOAD-FNet with this setup. For the survival prediction task, we evaluated the MOAD-FNet approach using the two most common baseline fusion models: ABMIL \cite{ilse2018attention} and TransMIL \cite{shao2021transmil}. We used the late fusion techniques: concatenation and Kronecker product and compared these against MOAB within both fusion settings.

\subsection{Data and Code Availability}
For brain tumor subtyping, we obtained data from NHNN through a rigorous application process to BRAIN UK, securing anonymized H\&E slides of tissue samples and epigenetic data. For survival prediction, WSIs are publicly available through the TCGA repository, with corresponding omic data from \cite{jaume2024modeling}. Source code will be made available upon acceptance.

\section{Results}

\textbf{Subtyping results. } In Tables \ref{tab1} and \ref{ablation}, we present the results for the subtyping task on the NHNN BRAIN UK dataset. MOAD-FNet integrated with ABMIL consistently demonstrates superior performance in brain tumor subtyping. Specifically, Table \ref{tab1} shows that the SNN model performs well across metrics using omics-only data, achieving an F1-Macro score of $0.726 (\pm 0.003)$, marginally outperforming the MLP. In contrast, WSI-only models show relatively low performance across metrics, with ABMIL achieving an F1-Macro of just $0.247 (\pm 0.004)$ and TransMIL performing slightly worse at $0.217 (\pm 0.019)$ likely due to the issue of indistinguishable patterns between most subtypes and the presence of underrepresented rare classes. These results indicate that omics data alone offers strong predictive power, however further improvements and interpretability are limited without incorporating WSI data. On the other hand, multimodal models substantially outperformed unimodal ones. For instance, ABMIL-MOAD-FNet achieved the best scores across all metrics, with an F1-Macro of $0.745 (\pm 0.025)$ and an F1-Micro of $0.820 (\pm 0.031)$. This represents an improvement of $0.027$ in F1-Macro compared to the second-best multimodal model, TransMIL-MOAD-FNet, further emphasizing the effectiveness of the MOAD-FNet architecture in leveraging multimodal data for richer and more informative representations. To assess statistical significance, we performed a Wilcoxon rank-sum test \cite{rainio2024evaluation} comparing the F1-Macro scores of ABMIL-MOAD-FNet with all other multimodal models in Table \ref{tab1}. The test yielded a $p$-value of $0.043$, indicating a statistically significant difference at the 0.05 significance level. These results demonstrate that integrating MOAD-FNet with advanced multimodal architectures leads to significant performance gains over single-stage fusion methods. This underscores the critical role of dual fusion strategies in effectively combining complementary features from multiple modalities, ultimately driving superior predictive accuracy and robustness.

\begin{table}[ht!]
\caption{\scriptsize{\textbf{Subtyping prediction Results} on the NHNN UK BRAIN dataset. We show MOAD-FNet combined with SOTA baseline models. The best performance is highlighted in \textbf{bold}. Concatenation and Kronecker products are denoted (Cat) and (KP). The gray rows correspond to models integrating MOAD-FNet or using the early fusion embeddings $\mathbf{p}_{ij}$. These demonstrate improved performance compared to standalone baseline models.}}\label{tab1}
\centering
\resizebox{1\columnwidth}{!}{
\begin{tabular}{lcccc}\hline
      Model &          F1-Macro & F1-Micro & Precision & Recall \\\hline
      \multicolumn{5}{c}{\textbf{Omics}} \\
      SNN  & $0.726$ \scriptsize{$(0.003)$} & $0.819$ \scriptsize{$(0.013)$} & $0.799$ \scriptsize{$(0.015)$} & $0.715$ \scriptsize{$(0.023)$} \\ 
      MLP  & $0.690$ \scriptsize{$(0.012)$} & $0.794$ \scriptsize{$(0.021)$} & $0.741$ \scriptsize{$(0.018)$} & $0.684$ \scriptsize{$(0.031)$} \\
      \hline
      \multicolumn{5}{c}{\textbf{WSI}} \\
      ABMIL & $0.247$ \scriptsize{$(0.004)$} & $0.442$ \scriptsize{$(0.004)$} & $0.280$ \scriptsize{$(0.022)$} & $0.026$ \scriptsize{$(0.250)$}  \\
      TransMIL & $0.217$ \scriptsize{$(0.019)$} & $0.434$ \scriptsize{$(0.003)$} & $0.240$ \scriptsize{$(0.024)$} & $0.230$ \scriptsize{$(0.013)$}  \\
      \hline
      \multicolumn{5}{c}{\textbf{Multimodal}} \\
      \rowcolor{lightgray}
      ACMIL MOAD-FNet & $0.501$ \scriptsize{$(0.011)$} & $0.729$ \scriptsize{$(0.031)$} & $0.602$ \scriptsize{$(0.032)$} & $0.490$ \scriptsize{$(0.014)$} \\
      TransMIL (Cat) & $0.711$ \scriptsize{$(0.005)$} & $0.799$ \scriptsize{$(0.005)$} & $0.744$ \scriptsize{$(0.003)$} & $0.707$ \scriptsize{$(0.002)$} \\
      TransMIL (KP) & $0.483$ \scriptsize{$(0.003)$} & $0.741$ \scriptsize{$(0.016)$} & $0.485$ \scriptsize{$(0.007)$} & $0.506$ \scriptsize{$(0.008)$} \\
      \rowcolor{lightgray} 
      TransMIL MOAD-FNet & $0.724$ \scriptsize{$(0.005)$} & $0.816$ \scriptsize{$(0.002)$} & $0.759$ \scriptsize{$(0.009)$} & $0.719$ \scriptsize{$(0.001)$} \\
      MCAT & $0.402$ \scriptsize{$(0.046)$} & $0.661$ \scriptsize{$(0.021)$} & $0.408$ \scriptsize{$(0.045)$} & $0.414$ \scriptsize{$(0.039)$} \\
      \rowcolor{lightgray}
      MCAT $p_{ij}$ & $0.432$ \scriptsize{$(0.015)$} & $0.703$ \scriptsize{$(0.017)$} & $0.441$ \scriptsize{$(0.031)$} & $0.446$ \scriptsize{$(0.011)$} \\
      SURVPATH & $0.424$ \scriptsize{$(0.008)$} & $0.697$ \scriptsize{$(0.012)$} & $0.478$ \scriptsize{$(0.038)$} & $0.425$ \scriptsize{$(0.013)$} \\
      \rowcolor{lightgray}
      SURVPATH $p_{ij}$ & $0.531$ \scriptsize{$(0.008)$} & $0.761$ \scriptsize{$(0.008)$} & $0.632$ \scriptsize{$(0.005)$} & $0.520$ \scriptsize{$(0.007)$} \\
      ABMIL (Cat) & $0.718$ \scriptsize{$(0.013)$} & $0.806$ \scriptsize{$(0.002)$} & $0.764$ \scriptsize{$(0.018)$} & $0.717$ \scriptsize{$(0.023)$} \\
      ABMIL (KP) & $0.447$ \scriptsize{$(0.024)$} & $0.737$ \scriptsize{$(0.020)$} & $0.481$ \scriptsize{$(0.043)$} & $0.464$ \scriptsize{$(0.032)$} \\
      \rowcolor{lightgray} 
      ABMIL MOAD-FNet & \textbf{0.745} \scriptsize{$(0.025)$} & \textbf{0.820} \scriptsize{$(0.013)$} & \textbf{0.769} \scriptsize{$(0.016)$} & \textbf{0.745} \scriptsize{$(0.035)$} \\
\hline
\end{tabular}}
\end{table}

\begin{figure*}[!t]
\centering
\includegraphics[width=0.7\textwidth]{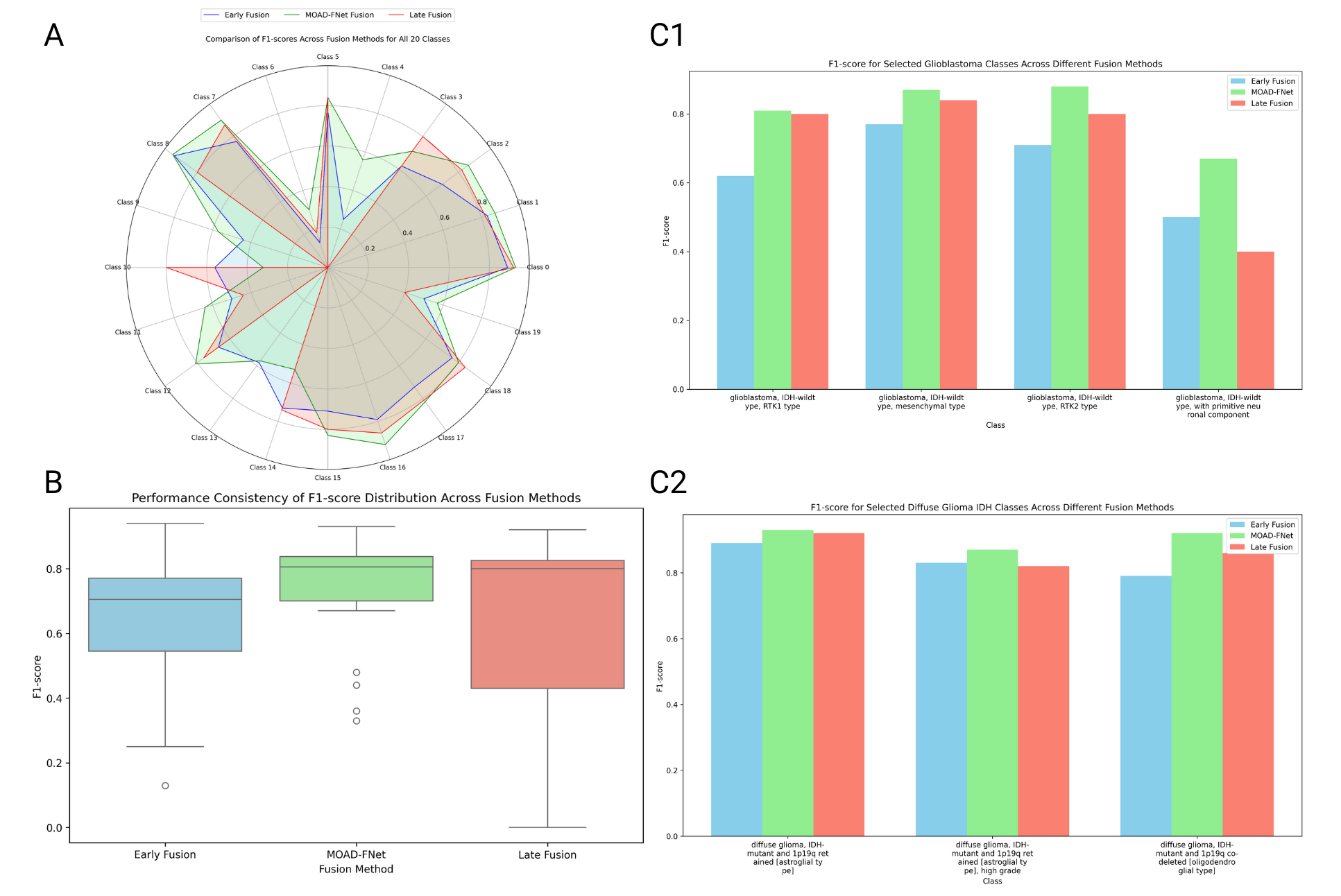}
\caption{Comparison of F1-scores across different fusion methods for glioma classification. The radar chart illustrates the F1-score performance across all 20 classes, highlighting distinct patterns for Early Fusion, MOAD-FNet, and Late Fusion. The bar plots zoom in on specific glioblastoma and glioma classes, showing class-level performance variations across fusion methods. The box plot provides a summary of F1-score distributions, showcasing the variability and consistency of each fusion method.}
\label{fig3}
\end{figure*}

\begin{figure*}[ht]
\centering
\includegraphics[scale=0.5]{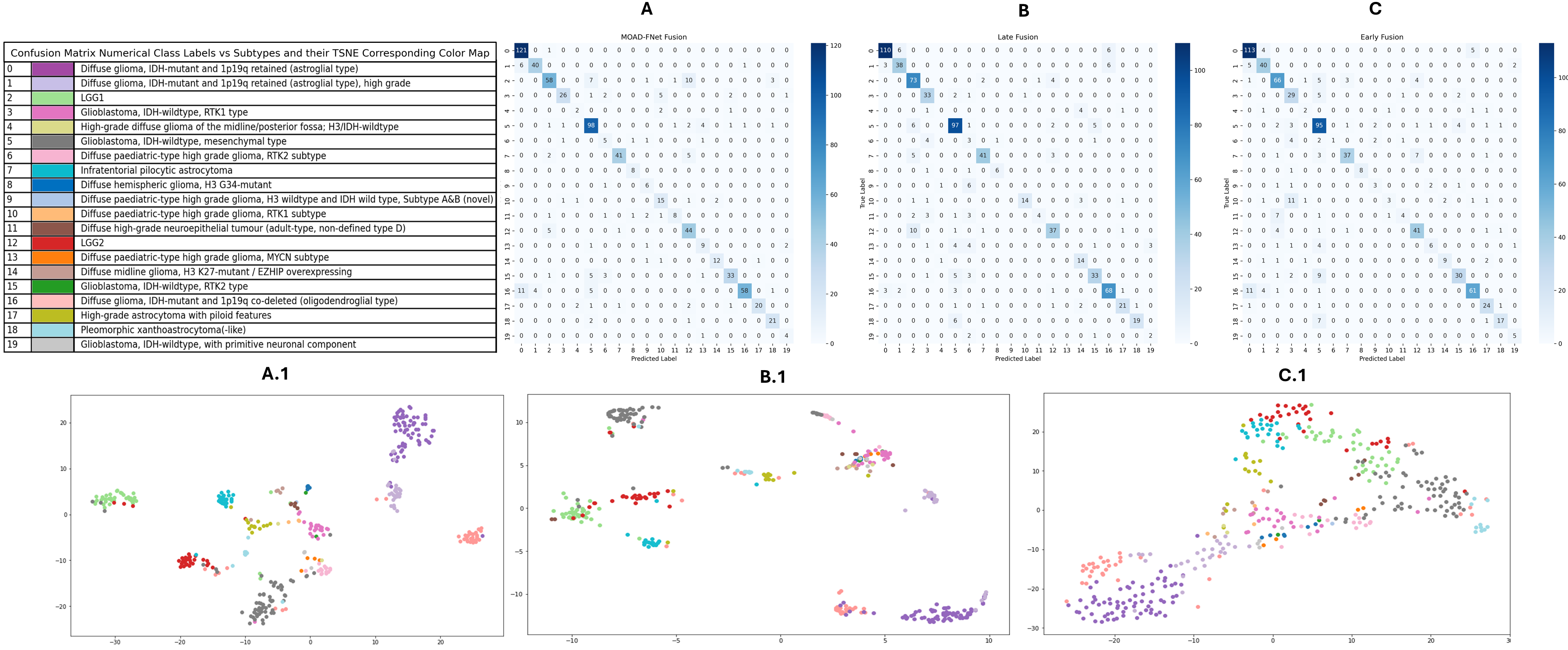}
\caption{Comparison of confusion matrices and t-SNE visualizations for three fusion strategies: (A) MOAD-FNet, (B) Late Fusion, and (C) Early Fusion for brain tumor subtyping. The corresponding t-SNE plots are labeled as (A.1), (B.1), and (C.1), respectively. }
\label{fig4}
\end{figure*}

Qualitative results shown in Fig. \ref{fig4} further illustrate MOAD-FNet’s effectiveness in classifying most tumor subtypes, showcasing the strong impact of its intermingled features. To assess class separation accuracy in the t-SNE representations we calculated silhouette scores for early and late fusion, as they displayed similar patterns. The late fusion t-SNE achieved a silhouette score of 0.33, while MOAD-FNet scored 0.37, indicating that it provides more accurate class separation. This is also evident in the box plot in Fig. \ref{fig3}, where MOAD-FNet displays fewer low outliers and a more compact distribution than early and late fusion. 

% ADD HEATMAP EXPLANATION

In Fig. \ref{fig4}A, we show the confusion matrix corresponding to our model MOAD-FNet, while in Fig. \ref{fig4}B and C, we see the results for late and early fusion, respectively. These demonstrate that our dual-fusion MOAD-FNet method excels in classifying subtypes, especially for minority classes. Notably, classes 4, 9, and 13, which was misclassified with late fusion (Fig. \ref{fig4}B), were correctly classified by MOAD-FNet. Conversely, late fusion performed better for classes 2 and 16, highlighting the challenges of handling heterogeneous multimodal data. This also underscores the need for tailored fusion techniques to address different data complexities. Furthermore, Fig. \ref{fig3}C1 and C2 illustrate that MOAD-FNet’s dual fusion approach considerably enhances classification of all glioblastoma and diffuse glioma subtypes, even in cases with highly overlapping morphological features. % This suggests that integrating multiple modalities provided critical discriminative information, greatly benefiting the classification of fine-grained subtypes.

We present ablation results in Table \ref{ablation}. The findings highlight the necessity for advanced fusion techniques to effectively integrate complementary features across modalities. Notably, the full MOAD-FNet model achieved the highest F1-Macro score of $0.745$ $(\pm0.035)$, demonstrating a notable improvement over both early fusion, with a gain of $0.101$, and late fusion, with a gain of $0.055$. Surprisingly, the task-agnostic ConvNeXt encoder achieved an F1-Macro of$ 0.732$ $(\pm0.012)$, marginally lower than the pretrained UNI encoder. This suggests that the core improvement in the full model stems not from the backbone architecture but from the dual fusion mechanism that effectively integrates complementary features across modalities. \\

%Placing the MOAB fusion block in the early fusion stage revealed two primary issues: first, element-wise multiplication in each outer arithmetic operation required reducing both feature dimensions to less than 64 to manage the high patch count (up to 10K per WSI), resulting in substantial information loss and suboptimal performance. Second, this approach demanded computational power and memory resources that exceeded feasible limits due to the large number of patches. \\

\begin{table}[h!]
\scriptsize
\caption{\scriptsize{\textbf{Subtyping prediction Ablation} study of MOAD-FNet showing the performance of early, late, and dual fusion methods using the NHNN BRAIN UK dataset.}}\label{ablation}
\centering
\begin{tabular}{c|p{2.5cm}|c}\hline
    Ablation & Model Description & F1-Macro \\ \hline
    ABMIL & Early fusion with $\mathbf{p}_{ij}$ & $0.644$ \scriptsize{$(0.010)$} \\
    ABMIL - MOAB & Late fusion with $\mathbf{e}_{ij}$ & $0.690$ \scriptsize{$(0.030)$} \\
    MOAD-FNet & ConvNeXt encoder & $0.732$ \scriptsize{$(0.012)$}\\
    \textbf{MOAD-FNet} & \textbf{Full model} & \textbf{$0.745$ \scriptsize{$(0.035)$}} \\ \hline
\end{tabular}
\end{table}

\textbf{Survival prediction results.} The results shown in Table \ref{tab3} demonstrate that integrating MOAD-FNet with existing SOTA methods (indicated by *) derives consistent performance improvement compared to baseline models. For instance, our approach achieves the highest c-index of $0.691 (\pm0.069)$ for BLCA and $0.726 (\pm0.049)$ for BRCA, surpassing all other methods in predicting patient disease-specific survival for BLCA while performing on par with the top-performing model MMP \cite{song2024multimodal} for BRCA. Interestingly, the best results on BRCA were obtained by PIBD with MOAB, achieving a strong c-index of $0.749 (\pm0.062)$ \cite{song2024multimodal}. % MMP and PIBD also show notable gains on BRCA compared to other approaches, with MMP achieving a c-index of 0.743 (±0.066).

\begin{table}[!ht]
\scriptsize
    \caption{\scriptsize{\textbf{Survival prediction Results} of MOAD-FNet with baselines for predicting patient disease-specific survival using the c-index. The best performance is highlighted in \textbf{bold}, and the second-best performance is underlined. Over five runs, the standard deviation is reported in brackets. Methods marked * are re-implemented.}}
    \label{tab3}
    \centering
    \begin{tabular}{lcc}
        \hline
        \textbf{Model} & \textbf{BLCA ($\uparrow$)} & \textbf{BRCA ($\uparrow$)} \\
        \hline
        \multicolumn{3}{c}{\textbf{WSI}} \\
        ABMIL*\cite{ilse2018attention}  & $0.572$ \scriptsize{$(0.084)$} & $0.573$ \scriptsize{$(0.097)$} \\
        TransMIL*\cite{shao2021transmil} & $0.579$ \scriptsize{$(0.052)$} & $0.611$ \scriptsize{$(0.011)$} \\ 
        \hline
        \multicolumn{3}{c}{\textbf{Omics}} \\
        MLP*  & $0.660$ \scriptsize{$(0.060)$} & $0.569$ \scriptsize{$(0.084)$} \\
        SNN* & $0.671$ \scriptsize{$(0.058)$} & $0.574$ \scriptsize{$(0.011)$} \\
        S-MLP* & $0.658$ \scriptsize{$(0.053)$} & $0.598$ \scriptsize{$(0.014)$} \\        
        \hline
        \multicolumn{3}{c}{\textbf{Multimodal}} \\
        PIBD \cite{zhang2024prototypical} & $0.667$ \scriptsize{$(0.061)$} & $0.736$ \scriptsize{$(0.072)$} \\
        PIBD* - MOAB & $0.684$ \scriptsize{$(0.046)$} & $\underline{0.749}$ \scriptsize{$(0.062)$} \\
        MMP \cite{song2024multimodal} & $0.628$ \scriptsize{$(0.064)$} & $\mathbf{0.753}$ \scriptsize{$(0.096)$} \\
        MMP \cite{song2024multimodal} & $0.635$ \scriptsize{$(0.064)$} & $0.738$ \scriptsize{$(0.096)$} \\
        ACMIL* - MOAD-FNet \cite{zhang2023attention} & $0.658$ \scriptsize{$(0.068)$} & $0.661$ \scriptsize{$(0.082)$} \\
        TransMIL* - MOAD-FNet & $0.661$ \scriptsize{$(0.053)$} & $0.675$ \scriptsize{$(0.068)$} \\
        SurvPath \cite{jaume2024modeling} & $0.625$ \scriptsize{$(0.056)$} & $0.655$ \scriptsize{$(0.089)$} \\
        SurvPath* - \textbf{$p_{ij}$} & $0.660$ \scriptsize{$(0.047)$} & $0.665$ \scriptsize{$(0.006)$} \\
        MBFusion \cite{zhang2024mbfusion} & $--$ & $0.644$ \scriptsize{$(0.020)$} \\
        ED-GNN \cite{ramanathan2024ensemble} & $--$ & $0.672$ \scriptsize{$(0.059)$} \\
        MoME \cite{xiong2024mome} & $\underline{0.686}$ \scriptsize{$(0.041)$} & $--$ \\
        MuGI \cite{long2024mugi} & $0.681$ \scriptsize{$(0.056)$} & $--$ \\
        \textbf{ABMIL MOAD-FNet(Ours)} & $\mathbf{0.691}$ \scriptsize{$(0.069)$} & $0.726$ \scriptsize{$(0.049)$} \\
        \hline
    \end{tabular}
\end{table}

Note that we did not test MOAD-FNet with MMP \cite{song2024morphological} because MMP already incorporates two early fusion stages (transformer and optimal transport). Adding MOAB would bring the total to four fusion stages, potentially introducing additional complexity and noise without clear performance benefits.

%Ablation studies demonstrate that MOAB performs best when combined with the ABMIL fusion baseline model using the Dual Fusion (DF) integration type. Additionally, DF surpasses all Late Fusion results in both the ABMIL and TransMIL  models
\begin{table}[!ht]
    \scriptsize
    \caption{\textbf{Survival prediction Ablation}. Comparison of c-index for TGGA BRCA and BLCA datasets using ABMIL and TransMIL models with two fusion stages, late (LF) and dual (DF), across three aggregation methods.}
    \label{tab4}
    \centering
    \begin{tabular}{c|c|c|c|c}
    \hline
        \multicolumn{5}{c}{\textbf{BRCA c-index ($\uparrow$)}}\\
        \hline
        \textbf{Model} & \textbf{Fus-S} & \textbf{Cat} & \textbf{KP} & \textbf{MOAB} \\
         TransMIL  & LF & $0.606$ \scriptsize{$(0.064)$} & $0.602$ \scriptsize{$(0.089)$} & $0.620$ \scriptsize{$(0.115)$} \\
                   & DF & $0.623$ \scriptsize{$(0.032)$} & $0.645$ \scriptsize{$(0.067)$} & $0.675$ \scriptsize{$(0.068)$} \\ 
        \hline
         ABMIL     & LF & $0.625$ \scriptsize{$(0.075)$} & $0.616$ \scriptsize{$(0.090)$} & $0.711$ \scriptsize{$(0.095)$}  \\
                   & DF & $0.634$ \scriptsize{$(0.023)$} & $0.640$ \scriptsize{$(0.058)$} & $\mathbf{0.726}$ \scriptsize{$(0.049)$} \\ 
        \hline
        \multicolumn{5}{c}{\textbf{BLCA c-index ($\uparrow$)}}\\
        \hline
        \textbf{Model} & \textbf{Fus-S} & \textbf{Cat} & \textbf{KP} & \textbf{MOAB} \\
         TransMIL  & LF & $0.599$ \scriptsize{$(0.081)$} & $0.582$ \scriptsize{$(0.053)$} & $0.600$ \scriptsize{$(0.108)$} \\
                   & DF & $0.624$ \scriptsize{$(0.068)$} & $0.595$ \scriptsize{$(0.052)$} & $0.661$ \scriptsize{$(0.053)$} \\
        \hline  
         ABMIL     & LF & $0.558$ \scriptsize{$(0.082)$} & $0.571$ \scriptsize{$(0.063)$} & $0.677$ \scriptsize{$(0.054)$}  \\
                   & DF & $0.622$ \scriptsize{$(0.063)$} & $0.561$ \scriptsize{$(0.054)$} & $\mathbf{0.691}$ \scriptsize{$(0.069)$} \\     
    \end{tabular}
\end{table}

%The results shown in Table \ref{tab3} demonstrate that integrating MOAD-FNet with existing SOTA methods (indicated by *) derives in consistent performance improvement compared to baseline models. Our approach obtains higher c-index scores across multiple methods, outperforming all other methods in predicting patient disease-specific survival for BLCA, and performing on par for BRCA. PIBD and MMP show notable performance gains on BRCA data compared to other approaches. To further explore this, we integrated MOAB within PIBD \cite{zhang2024prototypical}, achieving results comparable to the top performing model \cite{song2024multimodal}. However, we did not test MOAB with MMP \cite{song2024morphological} because MMP already incorporates two early fusion stages (transformer and optimal transport). Adding MOAB would bring the total to four fusion stages, potentially introducing additional complexity and noise without clear performance benefits.

% Discuss Ablation results here Table \ref{tab4}.
To further evaluate the effectiveness of MOAD-FNet in the survival prediction task, we conducted additional ablation studies presented in Table \ref{tab4}. The results demonstrate that the ABMIL model with Dual Fusion (DF) consistently delivers superior performance across both the BRCA and BLCA datasets, particularly when paired with the MOAB aggregation method. For the BRCA dataset, ABMIL-DF with MOAB achieves the highest c-index of $0.726 (\pm0.049)$, outperforming TransMIL-DF with MOAB. Similarly, for the BLCA dataset, ABMIL-DF with MOAB achieves the highest c-index of $0.691 (\pm0.069)$, surpassing TransMIL-DF with MOAB. The Late Fusion (LF) results follow a similar pattern, with ABMIL consistently outperforming TransMIL across all aggregation methods. These findings underscore the effectiveness of ABMIL, particularly with the Dual Fusion strategy and MOAB aggregation.

\textbf{Limitations.} While our method demonstrates promising performance across multiple tasks, it’s important to acknowledge some limitations. MOAD-FNet faces two primary challenges: first, balancing early and late fusion presents a trade-off; although MOAD-FNet leverages both stages to capture the strengths of each, finding the optimal balance can be complex, with some cases exhibiting performance variations depending on the fusion stage. Second, our late fusion block incorporating MOAB operates on latent space vectors derived from the early fusion stage, where omics features have already been blended. Consequently, identifying the specific CpG feature with the most profound impact on the outcome becomes challenging.
%with dual fusion stages, controlling redundant features from omic data becomes limited. These redundancies can potentially hinder the model's ability to accurately predict certain subtypes, as seen in the confusion matrix where classes 2 and 3 in MOAD-FNet score lower than the ablated models shown in Fig. \ref{fig4}B and C.

%Notably, MOAD-FNet’s integration not only boosts individual model performance but also shows consistent gains across diverse frameworks, such as TransMIL, SurvPath, and ACMIL, demonstrating its adaptability and broad applicability. This consistency reinforces MOAD-FNet’s robustness as a fusion framework that effectively captures the heterogeneous and complementary information within multimodal framework. Additionally, the ablation study results in Table \ref{tab4} provide further evidence of the superiority of the MOAD-FNet architecture, particularly when paired with the ABMIL baseline and the MOAB integration type.% These findings suggest that MOAD-FNet enhances survival prediction accuracy and offers a stable, high-performing multimodal solution that leverages integration strategies and advanced fusion techniques. 

% For the survival prediction task, these results shwo ...

\section{Conclusion}

The MOAD-FNet framework advances CNS tumor subtyping by effectively integrating DNA methylation and WSI data through a novel dual fusion approach. Our framework addresses the under-explored potential of combining these modalities, with the potential for distinguishing subtle morphological differences between tumor subtypes from DNA methylation profiling. The early fusion stage, implemented through an MLP-based mapping of WSI and methylation features, enables interpretable visualization while maintaining low dimensionality. The late fusion stage, enriched by outer arithmetic operations with MOAB, captures complex inter-modal relationships beyond simple addition. MOAD-FNet demonstrates superior performance across all evaluation metrics and exhibits robust scalability with different architectures. The framework's consistent success in both tumor subtyping and survival prediction establishes its practical utility for precision oncology. These results highlight how integration of multiple data modalities can enhance diagnostic accuracy while preserving clinical interpretability, thereby advancing automated approaches to CNS tumor classification.

\bibliographystyle{IEEEtran}
\bibliography{references.bib}

\begin{thebibliography}{10}
\providecommand{\url}[1]{#1}
\csname url@samestyle\endcsname
\providecommand{\newblock}{\relax}
\providecommand{\bibinfo}[2]{#2}
\providecommand{\BIBentrySTDinterwordspacing}{\spaceskip=0pt\relax}
\providecommand{\BIBentryALTinterwordstretchfactor}{4}
\providecommand{\BIBentryALTinterwordspacing}{\spaceskip=\fontdimen2\font plus
\BIBentryALTinterwordstretchfactor\fontdimen3\font minus \fontdimen4\font\relax}
\providecommand{\BIBforeignlanguage}[2]{{%
\expandafter\ifx\csname l@#1\endcsname\relax
\typeout{** WARNING: IEEEtran.bst: No hyphenation pattern has been}%
\typeout{** loaded for the language `#1'. Using the pattern for}%
\typeout{** the default language instead.}%
\else
\language=\csname l@#1\endcsname
\fi
#2}}
\providecommand{\BIBdecl}{\relax}
\BIBdecl

\bibitem{lopomo2018epigenetic}
A.~Lopomo and F.~Copped{\`e}, ``Epigenetic signatures in the diagnosis and prognosis of cancer,'' in \emph{Epigenetic Mechanisms in Cancer}.\hskip 1em plus 0.5em minus 0.4em\relax Elsevier, 2018, pp. 313--343.

\bibitem{liang2023integrative}
W.-W. Liang, R.~J.-H. Lu, R.~G. Jayasinghe, S.~M. Foltz, E.~Porta-Pardo, Y.~Geffen, M.~C. Wendl, R.~Lazcano, I.~Kolodziejczak, Y.~Song \emph{et~al.}, ``Integrative multi-omic cancer profiling reveals dna methylation patterns associated with therapeutic vulnerability and cell-of-origin,'' \emph{Cancer cell}, vol.~41, no.~9, pp. 1567--1585, 2023.

\bibitem{drexler2024unclassifiable}
R.~Drexler, F.~Brembach, J.~Sauvigny, F.~L. Ricklefs, A.~Eckhardt, H.~Bode, J.~Gempt, K.~Lamszus, M.~Westphal, U.~Sch{\"u}ller \emph{et~al.}, ``Unclassifiable cns tumors in dna methylation-based classification: clinical challenges and prognostic impact,'' \emph{Acta Neuropathologica Communications}, vol.~12, no.~1, p.~9, 2024.

\bibitem{djirackor2021intraoperative}
L.~Djirackor, S.~Halldorsson, P.~Niehusmann, H.~Leske, D.~Capper, L.~P. Kuschel, J.~Pahnke, B.~J. Due-T{\o}nnessen, I.~A. Langmoen, C.~J. Sandberg \emph{et~al.}, ``Intraoperative dna methylation classification of brain tumors impacts neurosurgical strategy,'' \emph{Neuro-Oncology Advances}, vol.~3, no.~1, p. vdab149, 2021.

\bibitem{pickles2020dna}
J.~C. Pickles, A.~R. Fairchild, T.~J. Stone, L.~Brownlee, A.~Merve, S.~A. Yasin, A.~Avery, S.~W. Ahmed, O.~Ogunbiyi, J.~G. Zapata \emph{et~al.}, ``Dna methylation-based profiling for paediatric cns tumour diagnosis and treatment: a population-based study,'' \emph{The lancet child \& adolescent health}, vol.~4, no.~2, pp. 121--130, 2020.

\bibitem{jaunmuktane2019methylation}
Z.~Jaunmuktane, D.~Capper, D.~T. Jones, D.~Schrimpf, M.~Sill, M.~Dutt, N.~Suraweera, S.~M. Pfister, A.~von Deimling, and S.~Brandner, ``Methylation array profiling of adult brain tumours: diagnostic outcomes in a large, single centre,'' \emph{Acta neuropathologica communications}, vol.~7, pp. 1--18, 2019.

\bibitem{smith2022major}
H.~L. Smith, N.~Wadhwani, and C.~Horbinski, ``Major features of the 2021 who classification of cns tumors,'' \emph{Neurotherapeutics}, vol.~19, no.~6, pp. 1691--1704, 2022.

\bibitem{chen2022pan}
R.~J. Chen, M.~Y. Lu, D.~F. Williamson, T.~Y. Chen, J.~Lipkova, Z.~Noor, M.~Shaban, M.~Shady, M.~Williams, B.~Joo \emph{et~al.}, ``Pan-cancer integrative histology-genomic analysis via multimodal deep learning,'' \emph{Cancer Cell}, vol.~40, no.~8, pp. 865--878, 2022.

\bibitem{capper2018dna}
D.~Capper, D.~T. Jones, M.~Sill, V.~Hovestadt, D.~Schrimpf, D.~Sturm, C.~Koelsche, F.~Sahm, L.~Chavez, D.~E. Reuss \emph{et~al.}, ``Dna methylation-based classification of central nervous system tumours,'' \emph{Nature}, vol. 555, no. 7697, pp. 469--474, 2018.

\bibitem{chen2021multimodal}
R.~J. Chen, M.~Y. Lu, W.-H. Weng, T.~Y. Chen, D.~F. Williamson, T.~Manz, M.~Shady, and F.~Mahmood, ``Multimodal co-attention transformer for survival prediction in gigapixel whole slide images,'' in \emph{Proceedings of the IEEE/CVF ICCV}, 2021, pp. 4015--4025.

\bibitem{chen2024towards}
R.~J. Chen, T.~Ding, M.~Y. Lu, D.~F. Williamson, G.~Jaume, A.~H. Song, B.~Chen, A.~Zhang, D.~Shao, M.~Shaban \emph{et~al.}, ``Towards a general-purpose foundation model for computational pathology,'' \emph{Nature Medicine}, vol.~30, no.~3, pp. 850--862, 2024.

\bibitem{hoang2024prediction}
D.-T. Hoang, E.~D. Shulman, R.~Turakulov, Z.~Abdullaev, O.~Singh, E.~M. Campagnolo, H.~Lalchungnunga, E.~A. Stone, M.~P. Nasrallah, E.~Ruppin \emph{et~al.}, ``Prediction of dna methylation-based tumor types from histopathology in central nervous system tumors with deep learning,'' \emph{Nature Medicine}, pp. 1--10, 2024.

\bibitem{jaume2024modeling}
G.~Jaume, A.~Vaidya, R.~J. Chen, D.~F. Williamson, P.~P. Liang, and F.~Mahmood, ``Modeling dense multimodal interactions between biological pathways and histology for survival prediction,'' in \emph{Proceedings of the IEEE/CVF Conference on Computer Vision and Pattern Recognition (CVPR)}, 2024, pp. 11\,579--11\,590.

\bibitem{capper2018practical}
D.~Capper, D.~Stichel, F.~Sahm, D.~T. Jones, D.~Schrimpf, M.~Sill, S.~Schmid, V.~Hovestadt, D.~E. Reuss, C.~Koelsche \emph{et~al.}, ``Practical implementation of dna methylation and copy-number-based cns tumor diagnostics: the heidelberg experience,'' \emph{Acta neuropathologica}, vol. 136, pp. 181--210, 2018.

\bibitem{hwang2024image}
J.~Hwang, Y.~Lee, S.-K. Yoo, and J.-I. Kim, ``Image-based deep learning model using dna methylation data predicts the origin of cancer of unknown primary,'' \emph{Neoplasia}, vol.~55, p. 101021, 2024.

\bibitem{zheng2020whole}
H.~Zheng, A.~Momeni, P.-L. Cedoz, H.~Vogel, and O.~Gevaert, ``Whole slide images reflect dna methylation patterns of human tumors,'' \emph{NPJ genomic medicine}, vol.~5, no.~1, p.~11, 2020.

\bibitem{sturm2023multiomic}
D.~Sturm, D.~Capper, F.~Andreiuolo, M.~Gessi, C.~K{\"o}lsche, A.~Reinhardt, P.~Sievers, A.~K. Wefers, A.~Ebrahimi, A.~K. Suwala \emph{et~al.}, ``Multiomic neuropathology improves diagnostic accuracy in pediatric neuro-oncology,'' \emph{Nature medicine}, vol.~29, no.~4, pp. 917--926, 2023.

\bibitem{mobadersany2018predicting}
P.~Mobadersany, S.~Yousefi, M.~Amgad, D.~A. Gutman, J.~S. Barnholtz-Sloan, J.~E. Vel{\'a}zquez~Vega, D.~J. Brat, and L.~A. Cooper, ``Predicting cancer outcomes from histology and genomics using convolutional networks,'' \emph{Proceedings of the National Academy of Sciences}, vol. 115, no.~13, pp. E2970--E2979, 2018.

\bibitem{zhang2024mbfusion}
Z.~Zhang, W.~Yin, S.~Wang, X.~Zheng, and S.~Dong, ``Mbfusion: Multi-modal balanced fusion and multi-task learning for cancer diagnosis and prognosis,'' \emph{Computers in Biology and Medicine}, vol. 181, p. 109042, 2024.

\bibitem{zhang2024prototypical}
Y.~Zhang, Y.~Xu, J.~Chen, F.~Xie, and H.~Chen, ``Prototypical information bottlenecking and disentangling for multimodal cancer survival prediction,'' \emph{ICLR}, 2024.

\bibitem{ogundipe2024deep}
O.~Ogundipe, Z.~Kurt, and W.~L. Woo, ``Deep neural networks integrating genomics and histopathological images for predicting stages and survival time-to-event in colon cancer,'' \emph{Plos one}, vol.~19, no.~9, p. e0305268, 2024.

\bibitem{song2024multimodal}
A.~H. Song, R.~J. Chen, G.~Jaume, A.~J. Vaidya, A.~S. Baras, and F.~Mahmood, ``Multimodal prototyping for cancer survival prediction,'' \emph{ICML}, 2024.

\bibitem{zhang2023attention}
Y.~Zhang, H.~Li, Y.~Sun, S.~Zheng, C.~Zhu, and L.~Yang, ``Attention-challenging multiple instance learning for whole slide image classification,'' \emph{ECCV}, 2024.

\bibitem{ramanathan2024ensemble}
V.~Ramanathan, P.~Pati, M.~McNeil, and A.~L. Martel, ``Ensemble of prior-guided expert graph models for survival prediction in digital pathology,'' in \emph{International Conference on Medical Image Computing and Computer Assisted Intervention (MICCAI)}.\hskip 1em plus 0.5em minus 0.4em\relax Springer, 2024, pp. 262--272.

\bibitem{liu2024agnostic}
H.~Liu, Y.~Shi, Y.~Xu, A.~Li, and M.~Wang, ``Agnostic-specific modality learning for cancer survival prediction from multiple data,'' \emph{IEEE Journal of Biomedical and Health Informatics}, 2024.

\bibitem{xu2023multimodal}
Y.~Xu and H.~Chen, ``Multimodal optimal transport-based co-attention transformer with global structure consistency for survival prediction,'' in \emph{Proceedings of the IEEE/CVF ICCV}, 2023, pp. 21\,241--21\,251.

\bibitem{song2024morphological}
A.~H. Song, R.~J. Chen, T.~Ding, D.~F. Williamson, G.~Jaume, and F.~Mahmood, ``Morphological prototyping for unsupervised slide representation learning in computational pathology,'' in \emph{Proceedings of the IEEE/CVF Conference on Computer Vision and Pattern Recognition (CVPR)}, 2024, pp. 11\,566--11\,578.

\bibitem{alwazzan2023moab}
O.~Alwazzan, A.~Khan, I.~Patras, and G.~Slabaugh, ``Moab: Multi-modal outer arithmetic block for fusion of histopathological images and genetic data for brain tumor grading,'' in \emph{International Symposium on Biomedical Imaging (ISBI)}.\hskip 1em plus 0.5em minus 0.4em\relax IEEE, 2023.

\bibitem{nicoll2022brain}
J.~A. Nicoll, T.~Bloom, A.~Clarke, D.~Boche, and D.~Hilton, ``Brain uk: Accessing nhs tissue archives for neuroscience research,'' \emph{Neuropathology and Applied Neurobiology}, vol.~48, no.~2, p. e12766, 2022.

\bibitem{qupath}
P.~Bankhead, M.~B. Loughrey, J.~A. Fern{\'a}ndez, Y.~Dombrowski, D.~G. McArt, P.~D. Dunne, S.~McQuaid, R.~T. Gray, L.~J. Murray, H.~G. Coleman \emph{et~al.}, ``Qupath: Open source software for digital pathology image analysis,'' \emph{Scientific reports}, vol.~7, no.~1, pp. 1--7, 2017.

\bibitem{orozco2018epigenetic}
J.~I. Orozco, T.~A. Knijnenburg, A.~O. Manughian-Peter, M.~P. Salomon, G.~Barkhoudarian, J.~R. Jalas, J.~S. Wilmott, P.~Hothi, X.~Wang, Y.~Takasumi \emph{et~al.}, ``Epigenetic profiling for the molecular classification of metastatic brain tumors,'' \emph{Nature communications}, vol.~9, no.~1, p. 4627, 2018.

\bibitem{gomes2022application}
R.~Gomes, N.~Paul, N.~He, A.~F. Huber, and R.~J. Jansen, ``Application of feature selection and deep learning for cancer prediction using dna methylation markers,'' \emph{Genes}, vol.~13, no.~9, p. 1557, 2022.

\bibitem{shao2021transmil}
Z.~Shao, H.~Bian, Y.~Chen, Y.~Wang, J.~Zhang, X.~Ji \emph{et~al.}, ``Transmil: Transformer based correlated multiple instance learning for whole slide image classification,'' \emph{Advances in Neural Information Processing systems}, vol.~34, pp. 2136--2147, 2021.

\bibitem{klambauer2017self}
G.~Klambauer, T.~Unterthiner, A.~Mayr, and S.~Hochreiter, ``Self-normalizing neural networks,'' \emph{Advances in NeurIPS}, vol.~30, 2017.

\bibitem{ilse2018attention}
M.~Ilse, J.~Tomczak, and M.~Welling, ``Attention-based deep multiple instance learning,'' in \emph{ICML}.\hskip 1em plus 0.5em minus 0.4em\relax PMLR, 2018, pp. 2127--2136.

\bibitem{jaume2024transcriptomics}
G.~Jaume, L.~Oldenburg, A.~Vaidya, R.~J. Chen, D.~F. Williamson, T.~Peeters, A.~H. Song, and F.~Mahmood, ``Transcriptomics-guided slide representation learning in computational pathology,'' in \emph{Proceedings of the IEEE/CVF Conference on Computer Vision and Pattern Recognition (CVPR)}, 2024, pp. 9632--9644.

\bibitem{liu2022convnet}
Z.~Liu, H.~Mao, C.-Y. Wu, C.~Feichtenhofer, T.~Darrell, and S.~Xie, ``A convnet for the 2020s,'' in \emph{Proceedings of the IEEE/CVF Conference on Computer Vision and Pattern Recognition (CVPR)}, 2022, pp. 11\,976--11\,986.

\bibitem{rainio2024evaluation}
O.~Rainio, J.~Teuho, and R.~Kl{\'e}n, ``Evaluation metrics and statistical tests for machine learning,'' \emph{Scientific Reports}, vol.~14, no.~1, p. 6086, 2024.

\bibitem{xiong2024mome}
C.~Xiong, H.~Chen, H.~Zheng, D.~Wei, Y.~Zheng, J.~J. Sung, and I.~King, ``Mome: Mixture of multimodal experts for cancer survival prediction,'' in \emph{International Conference on Medical Image Computing and Computer Assisted Intervention (MICCAI)}.\hskip 1em plus 0.5em minus 0.4em\relax Springer, 2024, pp. 318--328.

\bibitem{long2024mugi}
L.~Long, J.~Cui, P.~Zeng, Y.~Li, Y.~Liu, and Y.~Wang, ``Mugi: Multi-granularity interactions of heterogeneous biomedical data for survival prediction,'' in \emph{International Conference on Medical Image Computing and Computer Assisted Intervention (MICCAI)}.\hskip 1em plus 0.5em minus 0.4em\relax Springer, 2024, pp. 490--500.

\end{thebibliography}
\end{document}